\documentclass[letterpaper]{article}
\usepackage[draft]{aaai2026}
\usepackage{times}
\usepackage{helvet}
\usepackage{courier}
\usepackage[hyphens]{url}
\usepackage{graphicx}
\usepackage{natbib}
\usepackage{caption}
\usepackage{algorithm}
\usepackage{algorithmic}
\usepackage{newfloat}
\usepackage{listings}
\DeclareCaptionStyle{ruled}{labelfont=normalfont,labelsep=colon,strut=off}
\floatstyle{ruled}
\newfloat{listing}{tb}{lst}{}
\floatname{listing}{Listing}
\DeclareCaptionStyle{ruled}{labelfont=normalfont,labelsep=colon,strut=off}
\title{Critical Acclaim Orientation in Large Language Models:\\
       Evidence from Film Preference Elicitation}
\author{
    Jonghyun Jee\textsuperscript{\rm 1},
    Aaron Shaw\textsuperscript{\rm 1}
}
\affiliations{
    \textsuperscript{\rm 1}Department of Communication Studies,
    Northwestern University\\
    jonghyun@u.northwestern.edu
}

\begin{document}

\maketitle

\begin{abstract}
Large language models (LLMs) are trained on corpora that contain expressions of human judgment about films, books, music, and more. Yet whether LLMs systematically reproduce evaluative hierarchies remains unclear. Prior research on cultural bias in LLMs suggests competing expectations: models may mirror the popularity signals of internet texts, or may reproduce forms of prestige embedded in critical discourse. We probe this question through a study of film evaluations with eight models from four families (Anthropic, OpenAI, Alibaba, and Mistral), using a 200-film benchmark partitioned into critically acclaimed, commercially successful, and dual-legitimacy (critical acclaim + commercial success) films. Across 20,000 pairwise forced-choice comparisons per model analyzed with Bradley--Terry estimation, we observe a consistent critical acclaim orientation with all models: critically acclaimed yet commercially obscure films are selected over commercially successful yet critically unrecognized ones. This pattern grows with model scale within each family. In addition, nested OLS regression analyses show that evaluative orientation, public visibility, and popular reception distinctly help explain preferences. Adjusting for public visibility reverses the models' preference for dual-legitimacy films over critical acclaim-only films, while additionally accounting for popular reception attenuates much of the disadvantage of films with commercial success only. Finally, evaluative and recommendation-oriented prompt framings produce divergent rankings, suggesting that critical acclaim orientation may manifest indirectly in real-world LLM deployments.
\end{abstract}

\begin{links}
     \link{Code \& Data }{https://github.com/jonghyunjee/llm-film-preference}
     \link{Supplementary Materials }{https://doi.org/10.5281/zenodo.20826032}
\end{links}

\section{Introduction}

Large language models are, as Weatherby \citeyearpar{Weatherby_2025} argues, computational-cultural interfaces that have ingested human culture on a large scale and learned to operate within its structures. But training corpora are not evaluatively neutral. They contain decades of film criticism, canonical lists, and cultural discourse that reflect longstanding distinctions between critical consecration and commercial success. We investigate whether LLMs reproduce these distinctions in their outputs, and through what mechanisms.

Two competing predictions frame our inquiry. One follows from visibility. Commercially successful films typically dominate online discourse in terms of volume. Fan communities, news coverage, and social media discussions tend to feature contemporary blockbuster films more than, say, arthouse cinema from the 1960s. If LLM outputs primarily reflect patterns of exposure in training-relevant corpora, models would favor widely visible cultural objects. An alternative expectation follows from cultural prestige. Critical discourse, which tends to reflect the consolidated judgments and preferences of cultural elites, is also plausibly represented in models' training data. If evaluative distinctions are embedded in language in ways that persist beyond differences in public visibility, models would reproduce forms of prestige associated with critical acclaim even when those objects receive less public attention.

Prior research has characterized how models encode social bias and asymmetries along demographic \cite{Duan_et_al_2025, Salinas_et_al_2025, Seth_et_al_2025}, national and linguistic \cite{Naous_et_al_2024, Poole-Dayan_et_al_2024}, and political dimensions \cite{Motoki_et_al_2023, Waight_et_al_2026}. Cultural preference---the tendency to favor certain cultural objects over others---falls under a distinct analytical category. It is not self-evidently a bias, which is presumably why it has remained outside the scope of standard fairness audits. Yet cultural sociologists have long established that preference is something more than individual idiosyncrasy: it is structured along axes of prestige and legitimacy that map onto social hierarchies \cite{DiMaggio_Useem_1978, Bourdieu_1984, Bryson_1996, Peterson_Kern_1996}. Such hierarchies are already latent in the distributional structure of language: word embeddings trained on large corpora recover a cultural dimension aligned with Bourdieu's notion of distinction \cite{Kozlowski_et_al_2019}. Whether those hierarchies surface in the evaluative outputs of LLMs, when models are asked directly which cultural objects they \textit{prefer}, is what this study seeks to understand.

We investigate this question through a pairwise comparison study of films on eight models from four families (Anthropic, OpenAI, Alibaba, Mistral). Using a 200-film benchmark stratified across critically acclaimed, commercially successful, and jointly recognized films, we elicit 20,000 comparisons per model to estimate the latent comparative strength of each film.

Among all eight models, we find a consistent pattern we term \textit{critical acclaim orientation}: LLM outputs favor critically acclaimed cultural objects over commercially successful yet critically unrecognized counterparts. The effect intensifies with model scale within each family.

To investigate the sources of this evaluative orientation, we disambiguate several distinct signals in the data. First, we separate the \textit{volume} of online discourse from its \textit{evaluative valence}. Volume, which we operationalize as \textbf{public visibility}, refers to the amount of discourse surrounding a cultural object. Valence dictates the normative judgment attached to that object, which we further categorize into two modes of legitimation: \textbf{critical acclaim} (consecration by cultural institutions) and \textbf{popular reception} (mass audience appeal). By fitting a series of nested regression models that incorporate proxy measures of public visibility (IMDb vote counts) and popular reception (IMDb user ratings), we assess whether the observed evaluative patterns remain after accounting for these competing signals.

The results suggest that LLMs exhibit critical acclaim orientation that is associated with, but distinct from either public visibility or popular reception. These findings extend prior work on cultural bias in LLMs and contribute evidence of a novel dimension of asymmetric LLM outputs. Agnostic to the question of whether algorithmic systems possess internal cognitive states or ``taste'', their behavioral outputs skew towards critically consecrated works.

\section{Related Work}

\subsection{Evaluative Orientations in LLMs}

Research on LLM outputs has revealed systematic patterns across racial, gender, and socioeconomic dimensions \cite{Arzaghi_et_al_2024, Bommasani_Liang_2024, Nicolas_Caliskan_2025}. A parallel body of work has asked whether LLMs hold structured attitudes or values. While such orientations exist, measuring them is notoriously sensitive to evaluation design and prompt framing \cite{Ma_et_al_2024, Sachdeva_van_Nuenen_2025}. Furthermore, model scale interacts with these orientations in unexpected ways. Larger models exhibit more pronounced biases in certain domains, an effect attributed to the broader knowledge base accumulated during pretraining \cite{Melotte_et_al_2022}, though the direction and magnitude of these scale effects is mainly governed by underlying data composition \cite{Ali_et_al_2024}.

A related yet distinct body of work examines cultural bias in LLM-based recommendation systems. Fairness benchmarks applied to movie and music recommendation show that outputs differ depending on user attributes including gender, language, and cultural background, with some groups consistently receiving less equitable treatment \cite{Zhang_et_al_2023}. When demographic or contextual signals such as socioeconomic status are present, these asymmetries amplify---mainstream content is favored while non-traditional options are marginalized across diverse user groups \cite{Sakib_Das_2024}. Even in cold-start settings where user information is minimal, models reproduce gendered and cultural stereotypes on recommendation domains \cite{Andre_et_al_2025}. The fairness concern in this literature is predominantly about the treatment of differently situated \textit{users}. What this framing does not address is the evaluative orientation of the model itself when there is no user profile or recommendation context to anchor the output.

\subsection{Competing Logics: Critical Acclaim vs. Popular Reception}

Prior work suggests that both critical acclaim and popular reception leave recoverable traces in language. Kozlowski et al. \citeyearpar{Kozlowski_et_al_2019} demonstrate that cultural status hierarchies are embedded within distributional word representations: ``golf,'' for instance, aligns more closely with affluence, whereas ``boxing'' projects onto the opposite end of the dimension. This encoding of prestige extends to domain-specific corpora, from the socioeconomic markers hidden in restaurant menus \cite{Jurafsky_et_al_2016} to the genre-specific taste hierarchies of YouTube music reviews \cite{Airoldi_2024}. Given that both forms of legitimation generate distinctive linguistic footprints, an important question is which evaluative logic dominates in LLM outputs.

One might hypothesize that popular reception would gain the upper hand. Research on the demographic composition of internet text shows that outputs skew toward the viewpoints of the Western, educated, industrialized, rich, and democratic (WEIRD; \citealt{Henrich_et_al_2010}) populations \cite{Atari_et_al_2023, Abdurahman_et_al_2024, Zhou_et_al_2025}. Gillespie \citeyearpar{Gillespie_2024} documents analogous ``quiet normativities'' in LLM-generated narratives, where unmarked prompts reproduce class-marked defaults. Following this logic, a Marvel film that is marketed to and discussed by the dominant demographic groups should generate sufficient discursive volume to overwhelm esoteric cinema from the Global South in a head-to-head comparison. This inference presumes that discursive volume translates into elicited choice, yet alignment and elicitation framing intervene between corpus-level regularities and what surfaces at prompt time.

\subsection{Eliciting Preferences from LLMs}

Extracting evaluative orientations from LLMs poses a methodological problem distinct from auditing representational patterns. Standard alignment interventions train models to suppress explicit value judgments, neutralizing open-ended queries with hedged or evasive language \cite{Ouyang_et_al_2022, Wei_et_al_2023}. However, models trained to feign neutrality will nonetheless reveal biased associations when subjected to forced-choice comparisons \cite{Bai_et_al_2025}. That is to say, pairwise binary choices effectively bypass the alignment layer's refusal mechanisms by forcing a decision.

However, elicited signals may still be noisy or unstable across evaluation conditions. For instance, LLM cultural alignment scores exhibit high instability throughout presentation formats and prompt phrasings \cite{Khan_et_al_2025}. This instability is not domain-specific. Even minor perturbations such as switching response option order or negating a statement degrade response consistency \cite{Shu_et_al_2024}. Credible preference measurement therefore demands rigorous stability verification and aggregation mechanisms to transform noisy pairwise judgments into a stable hierarchy.

The Bradley--Terry model \citeyearpar{Bradley_Terry_1952} provides such a framework. Recent literature confirms that aggregating LLM pairwise comparisons via Bradley--Terry yields more reliable rankings than extracting ordinal labels directly \cite{Wu_et_al_2024}. This approach has also been validated across contexts, proving capable of measuring the ideological positions of European political parties \cite{Leo_et_al_2025} and mapping open-ended survey data onto stable preference orderings \cite{DiGiuseppe_Flynn_2026}. Taken together, these studies justify both the forced-choice elicitation strategy and the aggregation approach adopted in this study.

\section{Methods}

\subsection{Dataset}

To operationalize and assess the critical-commercial distinction, we construct a 200-film benchmark from two corpora. Critical acclaim is indexed by \emph{They Shoot Pictures, Don't They?} (TSPDT) Top 1,000\footnote{https://theyshootpictures.com/gf1000.htm}, which aggregates lists and ballots from decades of professional film discourse. Commercial success is indexed by Box Office Mojo's (BOM) all-time global revenue rankings\footnote{https://www.boxofficemojo.com/chart/ww\_top\_lifetime\_gross}, which record audience uptake through theater revenue.

Films are assigned to three mutually exclusive sets. Set~A (dual-legitimacy, $n = 40$) contains all films that appear in both corpora; Set~B (critical-only, $n = 80$) draws from TSPDT-only entries absent from the BOM list; Set~C (commercial-only, $n = 80$) draws from BOM-only entries absent from the TSPDT. Before sampling, TSPDT entries are filtered to exclude television productions and films less than 60 minutes.

Within Sets~B and C, films are selected via stratified sampling that enforces both language group and era. Set~A is the full intersection and is retained intact. Because Set~A's dual-acclaim films skew toward English-language cinema (37 of 40 films) released from 1980 onward (33 of 40), Set~B (drawn from the TSPDT-only top 500) is structured to balance this distribution. Set~B oversamples historical cinema, with 46 of its 80 films released prior to 1980, and enforces quotas across a global pool. Its final composition includes 18 linguistic categories, limiting English-language films to 12. While Set C preserves the modern box office's heavy skew toward post-2000 ($n=61$) and English-language ($n=69$) releases, it also enforces quotas to capture non-English blockbusters (e.g., Chinese $n=8$, Japanese $n=2$). The complete 200-film list and the cross-tabulations of the stratified sampling are reported in Appendix~A.

The three sets differ substantially on two film-level covariates, which we present here as context and return to below in analyses that probe explanatory mechanisms. Median IMDb user ratings follow a strict ordering: Set~A ($8.30$), Set~B ($7.81$), Set~C ($6.98$), with all pairwise differences significant ($F(2, 197) = 74.29$, $p < .001$). Median IMDb vote counts show an even stronger ordering: Set~A (1,242,591), Set~C (312,582), Set~B (26,432), with Set~B carrying roughly $47\times$ fewer votes than Set~A ($F(2, 197) = 90.28$, $p < .001$).
    
\subsection{Models}

Eight models from four families are evaluated: Claude Haiku 4.5 and Claude Sonnet 4.6 (Anthropic, US); GPT-5.4 Nano and GPT-5.4 (OpenAI, US); Qwen2.5-Turbo and Qwen2.5-Plus (Alibaba, China); and Mistral Small 3.2 and Mistral Large 3 (Mistral AI, France). Within each family, the two models represent a small and large capability tier.

\subsection{Elicitation Procedure}

Cultural preference is elicited through pairwise forced-choice comparisons. This design avoids direct ordinal extraction from known canonical lists \citep{Wu_et_al_2024}. Each comparison presents two films in the format ``\textit{Title} (Year)'' and asks the model to respond with ``A'' or ``B'' only. Presentation order is randomized on every query to mitigate position bias. Temperature is set to 0 for all models.

The default prompt wording used throughout the study is: \textit{``Which of these two films do you prefer? A: \{item\_a\}; B: \{item\_b\}. Respond with either `A' or `B' only.''}

\subsection{Comparison Design}

Pair selection proceeds in three adaptive sampling phases. Phase~1 prioritizes broad coverage and is configured to bring films toward a minimum of 15 comparisons. Phase~2 shifts to competitive sampling and pairs films with similar current Bradley--Terry strengths to stabilize the middle of the ranking \cite{Bradley_Terry_1952}. Phase~3 concentrates the remaining budget on rank boundaries of interest, particularly within the upper quartile of the distribution.

This adaptive procedure trades uniform comparison density for estimation efficiency. In a Bradley--Terry framework, once provisional hierarchies emerge, pairings between films of widely different latent strengths yield minimal information gain. The design increases the informational value of each query by allocating more comparison budget to closely matched items and the upper quartile in Phases 2 and 3. Consequently, an endogenous exposure effect is introduced, as films that perform well early on naturally accrue more comparisons. Whereas this non-uniformity risks path-dependence, our cross-iteration stability checks (H1b) address this concern. If adaptive fluctuations drove rankings, inter-run Spearman $\rho$ would be low. The high observed $\rho$ across all eight models suggests that adaptive sampling converges to stable estimates.

Each model is queried via API for 4,000 comparisons per iteration over five independent runs, yielding 20,000 total comparisons per model and 160,000 across all eight models. The Bradley--Terry model assigns each film $i$ a latent strength parameter $\pi_i > 0$, such that the probability that film $i$ is preferred over film $j$ is

\begin{equation}
    P(i \succ j) = \frac{\pi_i}{\pi_i + \pi_j}.
    \label{eq:bt}
\end{equation}

Parameters are estimated via a custom implementation of the MM algorithm \cite{Hunter_2004}, which iterates the update

\begin{equation}
    \pi_i^{(t+1)} = \frac{w_i^+}{\displaystyle\sum_{j \neq i} \frac{n_{ij}}{\pi_i^{(t)} + \pi_j^{(t)}}},
    \label{eq:mm}
\end{equation}

where $w_i^+$ is the regularized win total of item $i$ (Dirichlet-$\frac{1}{2}$ pseudo-counts prevent divergence for undefeated or winless items) and $n_{ij}$ is the total number of comparisons between $i$ and $j$, until convergence ($\max_i |\Delta\pi_i / \pi_i| < 10^{-10}$, up to 2{,}000 iterations). Log-strengths $\lambda_i = \log \pi_i$ are mean-centered after each run. Standard errors are computed from the full inverse Fisher information matrix, accounting for covariance between items with shared opponents. Final $\lambda_i$ for each film are averaged across five independent iterations; within-model values are then $z$-scored to produce $\lambda_z$.

\subsection{Validation of elicited preferences (H1)}

A first hypothesis set assesses the validity of model preferences. Validation conditions as well as pass criteria, described below, were specified for three sub-hypotheses (H1a--H1c) before analysis and are consistent with established benchmarks for acceptable measurement reliability \cite{Schober_Boer_Schwarte_2018}. We also conducted an additional diagnostic (H1d) to evaluate the structural coherence of the elicited preference graph and the adequacy of the Bradley--Terry model used in subsequent analyses.

\paragraph{H1a: Item-level determinism.} At temperature~0, the same pairwise comparison is presented 10 times to each model, randomly drawn from a sample of film pairs. Pass criterion: mean within-pair consistency $\geq 0.95$ per model. Consistency is computed as the proportion of responses that match the majority answer; a value of 1.0 indicates perfect determinism and 0.5 indicates chance-level flipping.

\paragraph{H1b: Ranking stability across iterations.} All pairwise Spearman $\rho$ between per-run Bradley--Terry ranking vectors are computed within each model over five independent iterations. Pass criterion: mean $\rho > 0.80$.

\paragraph{H1c: Prompt frame invariance.} Four semantically similar but lexically distinct preference wordings are administered to a subset of four large-tier models at 4,000 comparisons per wording, yielding one Bradley--Terry ranking vector per wording per model:

\begin{enumerate}
    \item \textit{``Which of these two films do you prefer?''} (baseline)
    \item \textit{``Which of these two films do you like more?''}
    \item \textit{``Which of these two films is closer to your taste?''}
    \item \textit{``You are recommending a film to a general audience. Which would you recommend?''}
\end{enumerate}

The fourth framing serves as a contrastive condition expected to produce divergent rankings by shifting from evaluative to audience-fit language. Spearman $\rho$ is computed between all wording pairs within each model. Pass criterion: mean $\rho > 0.80$ over the three evaluative wordings (1--3).

\paragraph{H1d: Structural coherence and model adequacy.} We also evaluate preference transitivity as well as the predictive performance of fitted values of $\lambda_i$.  First, we enumerate all film triads whose three pairs were observed and calculate the proportion that form directed cycles (weak stochastic transitivity, or WST, violations), compared against null models that preserve the observed comparison graph and per-pair comparison counts: a fair-coin tournament and a Bradley--Terry null drawn from each model's fitted strengths. Then, we assess the adequacy of the Bradley-Terry model using five-fold cross-validation over held-out pairs to measure how well the fitted strengths predict unseen comparisons.

\subsection{Analysis (H2--H4)}

Three hypotheses guide the analyses. \textbf{H2} predicts that critically acclaimed films (Set~B) will win significantly more than 50\% of head-to-head matchups against commercially successful films (Set~C). \textbf{H3} predicts that dual-legitimacy films (Set~A) will defeat critically acclaimed-only films (Set~B) in direct comparisons. \textbf{H4} predicts that larger models within the same family will exhibit a stronger critical acclaim orientation than smaller models, operationalized as a higher Set~B vs.\ Set~C win rate.

\paragraph{Primary analysis.} For each model, the proportion of Set~B vs. Set~C head-to-head comparisons won by Set~B is tested against a 0.50 null via binomial test \textbf{(H2)}. Set~A win rates against both B and C are reported to assess dual-legitimacy predictions \textbf{(H3)}. We use Mann--Whitney $U$ as the non-parametric test statistic for both and report effect sizes as Cohen's $d$ on standardized $\lambda_z$ distributions.

\paragraph{Scale analysis.} Within-family comparisons of B~vs.~C win rates between small and large models test whether critical acclaim orientation intensifies with model scale \textbf{(H4)}. Smaller models with a thinner representation of esoteric cinema may default to commercially successful films (Set~C). To test this knowledge coverage mechanism, the A~vs.~C win rate is compared against the B~vs.~C win rate by tier. If the A~vs.~C rate remains stable while the B~vs.~C rate rises with scale, it suggests that the model is partially overcoming initial ignorance.

\paragraph{Regression analysis.} Bradley--Terry log-strengths ($\lambda_i$) are standardized (z-scored) within each model before pooling, yielding an approximately Gaussian outcome distribution ($N = 1{,}600$ film$\times$model observations). Because this within-model standardization centers each model's distribution at zero, it reduces model-level intercept variance and the primary source of intra-cluster correlation. Consequently, pooled Ordinary Least Squares (OLS) with robust standard errors is used. A series of four nested models (M1--M4) regresses standardized $\lambda_z$ on film-level predictors. M1 includes corpus set membership only. M2 adds era (pre-1960, 1960s--70s, 1980s--90s, 2000s+). M3 adds public visibility (proxied by log IMDb votes). M4 adds popular reception (proxied by median IMDb user ratings). Models are compared via sequential F-tests. Descriptive statistics for all regression variables are reported in Appendix Table~A3.

\begin{table*}[t]
\centering
\caption{Win rates and effect sizes for Set~B vs.\ C, Set~A vs.\ B, and Set~A vs.\ C.}
\label{tab:results}
\small
\begin{tabular}{lllrrrrrr}
\hline
\hline
 & & & \multicolumn{2}{c}{B vs.\ C (H2 \& H4)} & \multicolumn{2}{c}{A vs.\ B (H3)} & \multicolumn{2}{c}{A vs.\ C (H4)} \\
\cline{4-5} \cline{6-7} \cline{8-9}
Family & Model & Size & Win \% & $d$ & Win \% & $d$ & Win \% & $d$ \\
\hline
OpenAI
 & GPT-5.4           & Large & $87.8\%^{***}$ & $1.44$ & $52.7\%^{***}$ & $+0.23$ & $90.2\%^{***}$ & $+1.62$ \\
 & GPT-5.4 Nano      & Small & $70.3\%^{***}$ & $0.82$ & $60.8\%^{***}$ & $+0.72$ & $89.5\%^{***}$ & $+1.68$ \\[3pt]
Anthropic
 & Claude Sonnet~4.6 & Large & $77.7\%^{***}$ & $1.31$ & $48.2\%$ n.s.  & $+0.22$ & $91.4\%^{***}$ & $+1.60$ \\
 & Claude Haiku~4.5  & Small & $65.6\%^{***}$ & $0.97$ & $50.6\%$ n.s.  & $+0.46$ & $79.9\%^{***}$ & $+1.37$ \\[3pt]
Mistral
 & Mistral Large~3   & Large & $84.1\%^{***}$ & $1.72$ & $41.2\%$ n.s.  & $-0.27$ & $89.0\%^{***}$ & $+1.48$ \\
 & Mistral Small~3.2 & Small & $77.1\%^{***}$ & $1.11$ & $58.6\%^{***}$ & $+0.62$ & $89.7\%^{***}$ & $+1.80$ \\[3pt]
Alibaba
 & Qwen2.5-Plus      & Large & $79.7\%^{***}$ & $1.32$ & $41.9\%$ n.s.  & $-0.15$ & $85.2\%^{***}$ & $+1.26$ \\
 & Qwen2.5-Turbo     & Small & $66.3\%^{***}$ & $0.90$ & $46.3\%$ n.s.  & $+0.08$ & $76.6\%^{***}$ & $+1.00$ \\
\hline
\end{tabular}
\\[4pt]
\raggedright\small
Binomial test against $p = 0.50$. Unmarked win rates are non-significant (n.s.). Effect sizes ($d$) are computed on standardized $\lambda_z$ distributions via Mann--Whitney $U$; For B~vs.\ C: $d > 0$ indicates Set~B preferred. For A~vs.\ B and A~vs.\ C: positive $d$ = Set~A preferred over the named comparator; negative $d$ = comparator preferred over Set~A. $^{***}p < .001$.
\end{table*}

\section{Results}

%% -------------------------------------------------------
\subsection{H1: Validation of Elicited Preferences}
%% -------------------------------------------------------

The three pre-specified conditions were satisfied prior to substantive analysis. Full results are reported in Appendix~B.

\textbf{H1a (Item-level determinism).} At temperature~$= 0$, all eight models exceeded the 0.95 pass criterion (mean consistency range: 0.959--1.000). Three models (Claude Haiku~4.5, Qwen2.5-Plus, and Qwen2.5-Turbo) were perfectly deterministic in all 400 calls. Non-deterministic responses were concentrated in Set~A vs.\ Set~B comparisons where the $\lambda$-gap is small.

\textbf{H1b (Ranking stability across iterations).} Mean inter-run Spearman $\rho$ ranged from 0.831 (Claude Haiku~4.5) to 0.925 (GPT-5.4 Nano) among the models, all exceeding the 0.80 pass criterion.

\textbf{H1c (Prompt frame invariance).} All four large-tier models exceeded the mean $\rho > 0.80$ pass criterion across evaluative prompt pairs (range: 0.843--0.884), thereby supporting the stability of BT rankings over semantically similar preference wordings. The contrastive general-audience recommendation framing produced divergent rankings (mean GA $\rho \approx 0.26$--$0.31$ vs.\ evaluative mean $\rho \approx 0.85$), confirming that evaluative preference elicits a signal distinct from audience-fit recommendation. The baseline preference wording is adopted for all main analyses.

\textbf{H1d (Structural coherence).} The elicited preference graphs exhibit strong global coherence. Directed cycles are uncommon, with WST violation rates ranging from 3.8--12.7\%, well below both the fair-coin expectation of 25.0\% and the Bradley--Terry null (16.1--19.9\%) in every model. Analogous conditions on the magnitudes of comparison probabilities (moderate and strong stochastic transitivity) show the same pattern (see Appendix~B4). These results indicate that the elicited comparisons are more internally coherent than assumed by the standard Bradley--Terry observation model, suggesting that Bradley--Terry provides a conservative statistical approximation rather than imposing structure on noisy data. Consistent with this interpretation, five-fold cross-validation achieves held-out accuracies of 0.689--0.830, recovering 82--90\% of the maximum accuracy implied by each model's response consistency.

%% -------------------------------------------------------
\subsection{H2: Critical Acclaim Orientation}
%% -------------------------------------------------------

\begin{figure*}[ht]
\centering
\includegraphics[width=0.8\textwidth]{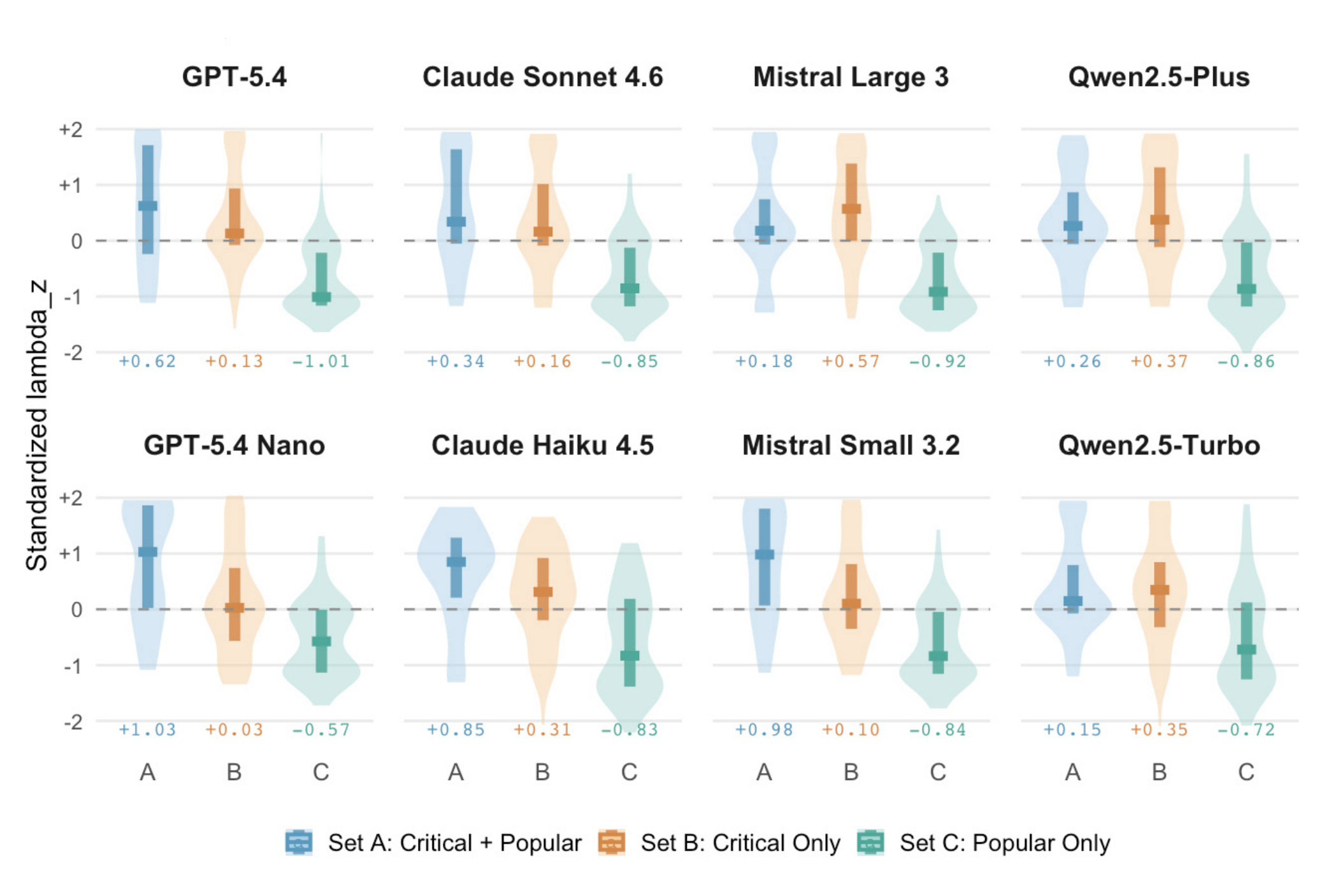}
\caption{$\lambda_z$ distributions by corpus set and model (H2, H3, H4). Each panel shows the full distribution of standardized preference strength for Sets~A, B, and~C. For each model-set pair, darker bars show the interquartile range and median values, while lighter shaded area visualizes the density. The per-set median is presented along the bottom of each panel. The consistent B~$>$~C separation in all panels supports H2; the mixed A vs. B pattern among models fails to reject the null for H3. A within-family comparison illustrates the scale effect (H4).}
\label{fig1}
\end{figure*}

H2 is supported in all eight models (Table~\ref{tab:results}). Set~B films (critically acclaimed, commercially obscure) won considerably more than 50\% of head-to-head matchups against Set~C films (commercially successful, critically unrecognized) (all $p < .001$).

Among large-tier models, Set~B win rates ranged from 77.7\% (Claude Sonnet~4.6) to 87.8\% (GPT-5.4), with effect sizes in the very large range ($d = 1.31$--$1.72$). Among small-tier models, win rates ranged from 65.6\% (Claude Haiku~4.5) to 77.1\% (Mistral Small~3.2), with $d = 0.82$--$1.11$. The distribution of standardized $\lambda_z$ scores for all three sets is shown in Figure~\ref{fig1}. Full per-model Bradley--Terry rankings for all 200 films are reported in Appendix~C.

The highest-ranked films among all models are \textit{Spirited Away} (2001, Japan), \textit{The Godfather} (1972, USA), \textit{Portrait of a Lady on Fire} (2019, France), \textit{Pather Panchali} (1955, India), and \textit{Yojimbo} (1961, Japan). The first two are Set~A dual-legitimacy films, while the rest are from Set~B, critically acclaimed but commercially obscure films. At the bottom are \textit{The Angry Birds Movie} (2016, USA), \textit{Twenty Years Later} (1984, Brazil), \textit{Cars 2} (2011, USA), \textit{The Fantastic Four: First Steps} (2025, USA), and \textit{Pegasus 2} (2024, China). With the exception of \textit{Twenty Years Later}, a Set~B film universally penalized, these lowest-ranked entries are Set~C films that occupy the bottom despite their commercial achievement. Full aggregated rankings are reported in Appendix Table~C9.

%% -------------------------------------------------------
\subsection{H3: Dual Legitimacy}
%% -------------------------------------------------------

H3, that Set~A (dual-legitimacy) films would defeat Set~B in head-to-head comparisons, is not supported. Among large-tier models, three of four show Set~B winning over Set~A: Mistral Large~3 (A win rate 41.2\%), Qwen~2.5 Plus (41.9\%), and Claude Sonnet~4.6 (48.2\%), none reaching statistical significance. Only GPT-5.4 shows a modest Set~A advantage (52.7\%, $p < .001$, $d = +0.23$). The pattern inverts among small-tier models, where three of four show Set~A winning significantly, but directional inconsistency across tiers precludes a coherent conclusion. For context, Set~A defeats Set~C across all eight models (win rates 76.6--91.4\%, all $p < .001$), affirming that dual legitimacy gives a strong preference advantage over pure commercial success.

%% -------------------------------------------------------
\subsection{H4: Scale-Dependent Critical Acclaim Orientation}
%% -------------------------------------------------------

H4 is supported. Within every family, the large-tier model exhibits a higher Set~B vs.\ Set~C win rate than the small-tier model ($\Delta$ win rate: $+7.1$ to $+17.5$ percentage points; $\Delta d$: $+0.34$ to $+0.62$; Table~\ref{tab:results}).

To probe the mechanism behind this effect, we compare how scale operates on two comparison types: Set~B vs.\ C, which contrasts critical only films against commercial only ones, and Set~A vs.\ C, which contrasts well-known dual-legitimacy films, likely densely represented in any training corpus at any scale, against the same commercial baseline. If scale acted primarily by improving coverage of obscure films, the B vs.\ C rate should rise while A vs.\ C remains flat.

The pattern of results for these comparisons is mixed. For OpenAI, the B~vs.\ C win rate rises by $+17.5$~pp with scale while the A~vs.\ C rate rises by only $+0.7$~pp; for Mistral, the corresponding figures are $+7.1$~pp vs.\ $-0.7$~pp. For Anthropic and Alibaba, however, both A~vs.\ C and B~vs.\ C win rates scale with model size at similar rates ($+11.5$~pp vs.\ $+12.1$~pp for Anthropic; $+8.5$~pp vs.\ $+13.4$~pp for Alibaba). Scalar increases of training data coverage alone cannot explain the differential contrasts. Set~A films are expected to be well-known at all scales, so their improved win rates against Set~C with larger models suggest a broader capability effect, even as the extent of that effect seems uneven.

\begin{table*}[t]
\centering
\caption{Nested OLS models predicting standardized BT log-strength ($\lambda_z$).}
\label{tab:regression}
\small
\begin{tabular}{lrrrr}
\hline
\hline
 & M1 & M2 & M3 & M4 \\
\hline
\textbf{Set B}
  & $-0.219^{***}$ & $-0.138^{*}$   & $+0.638^{***}$ & $+0.554^{***}$ \\
  & $(0.064)$      & $(0.068)$      & $(0.077)$      & $(0.068)$      \\[4pt]
\textbf{Set C}
  & $-1.192^{***}$ & $-1.136^{***}$ & $-0.801^{***}$ & $-0.150^{*}$   \\
  & $(0.061)$      & $(0.064)$      & $(0.065)$      & $(0.060)$      \\[4pt]
Era: 1960s--70s
  &                & $+0.429^{***}$ & $+0.379^{***}$ & $+0.425^{***}$ \\
  &                & $(0.091)$      & $(0.081)$      & $(0.073)$      \\
Era: 1980s--90s
  &                & $+0.617^{***}$ & $+0.392^{***}$ & $+0.421^{***}$ \\
  &                & $(0.089)$      & $(0.082)$      & $(0.076)$      \\
Era: 2000s$+$
  &                & $+0.350^{***}$ & $+0.236^{**}$  & $+0.480^{***}$ \\
  &                & $(0.095)$      & $(0.086)$      & $(0.078)$      \\[4pt]
log IMDb votes
  &                &                & $+0.235^{***}$ & $+0.094^{***}$ \\
  &                &                & $(0.014)$      & $(0.016)$      \\[4pt]
IMDb rating
  &                &                &                & $+0.714^{***}$ \\
  &                &                &                & $(0.032)$      \\
\hline
$N$                    & 1,600  & 1,600  & 1,600  & 1,600  \\
Adj.\ $R^{2}$          & 0.270  & 0.300  & 0.405  & 0.540  \\
$\Delta F$             & ---    & $36.53^{***}$ & $365.93^{***}$ & $467.98^{***}$ \\
\hline
\hline
\end{tabular}
\\[6pt]
\raggedright\small
Pooled OLS; outcome = standardized BT log-strength ($\lambda_z$); Set~A = reference category; pre-1960 = era reference category. Heteroscedasticity-consistent (HC3) standard errors in parentheses. $\Delta F$ = incremental $F$-test vs.\ prior model (M1$\to$M2: $F(3, 1594)$; M2$\to$M3: $F(1, 1593)$; M3$\to$M4: $F(1, 1592)$). $^{*}p < .05$, $^{**}p < .01$, $^{***}p < .001$.
\end{table*}

%% -------------------------------------------------------
\subsection{Modeling Preference Variations}
%% -------------------------------------------------------

Table~2 reports fixed-effect estimates across all models of standardized Bradley--Terry log strength. The base model (M1) contrasts the three sets of films with no covariates. To assess whether critical acclaim orientation gaps between sets reflect evaluative orientation or film-level confounds, each subsequent model introduces additional predictors: era (M2), a proxy for public visibility (log IMDb votes in M3), and a proxy for popular reception (IMDb user rating in M4).
 
\begin{figure*}[t]
\centering
\includegraphics[width=0.8\textwidth]{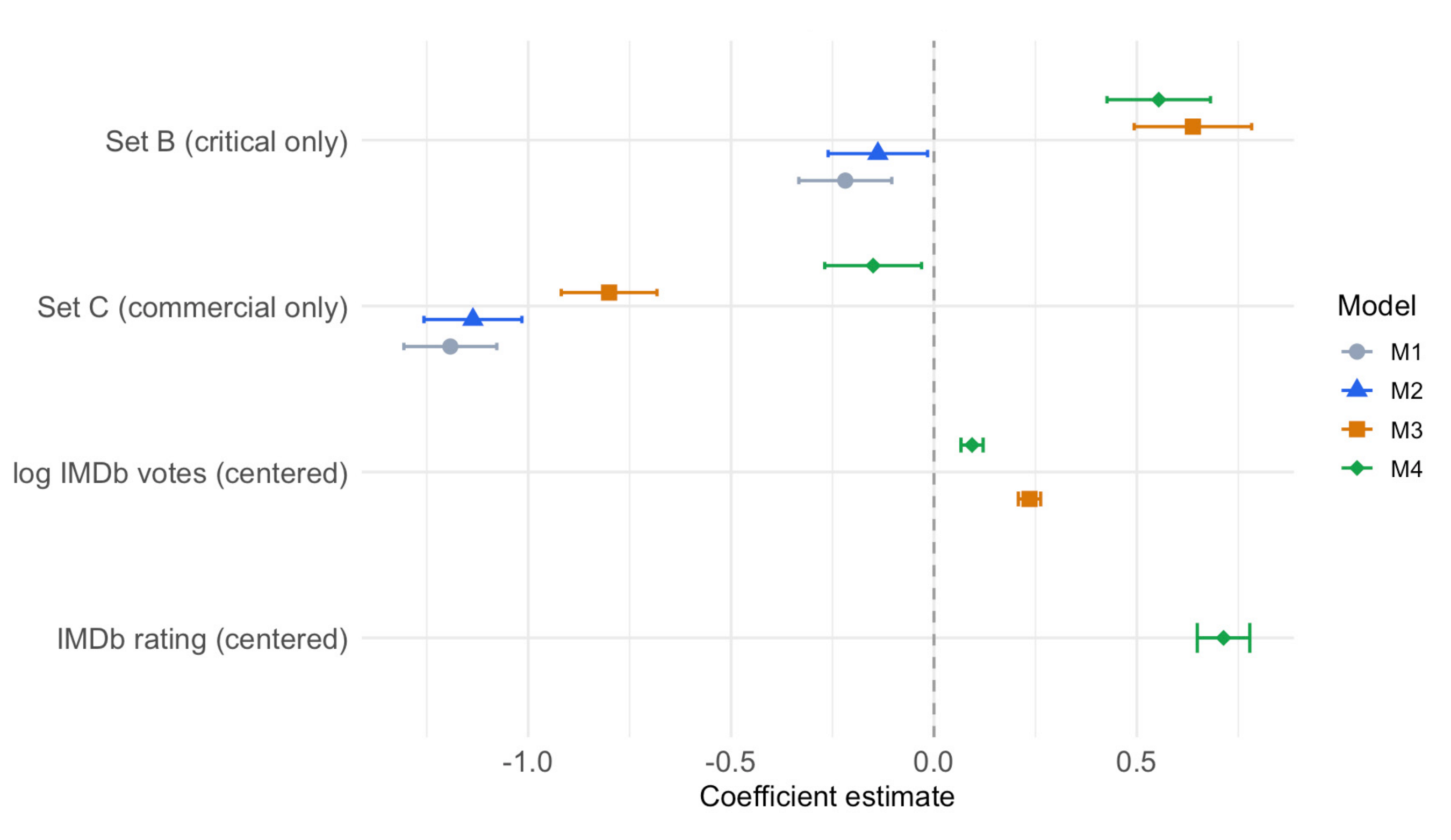}
\caption{Fixed-effect coefficients from nested OLS models M1--M4 (outcome: standardized $\lambda_z$; Set~A (dual legitimacy) = reference; 95\% CI). The sign reversal on Set~B (negative in M1 \& M2, positive in M3 \& M4 with the addition of public visibility and popular reception proxies) suggests that the preference disadvantage of critically acclaimed yet commercially obscure films reflects coverage asymmetry rather than evaluative dispreference. The progressive attenuation of the Set~C coefficient reflects the compression of the commercial penalty once public visibility and popular reception are held constant.}
\label{fig2}
\end{figure*}

\textbf{Set~B sign reversal.} M1 includes corpus set membership only, with Set~A as the reference category. Set~B films score  below Set~A ($b = -0.219$, $p < .001$), and Set~C films score substantially lower still ($b = -1.192$, $p < .001$). M2 adds era as a control; the Set~B coefficient remains negative ($b = -0.138$, $p = .027$). M3 adds log IMDb votes and the Set~B coefficient reverses sign ($b = +0.638$, $p < .001$). M4, the full model, adds IMDb user rating, yielding a Set~B estimate of $b = +0.554$ ($p < .001$). This sign reversal, alongside a substantial improvement in model goodness-of-fit, indicates that public visibility and popular reception explain much of the baseline variation in observed BT score differences between Set A and Set B. \textit{Ceteris paribus}, pure critical acclaim is associated with higher preference strength relative to dual-legitimacy films.

\textbf{Set~C attenuation.} Set~C's raw disadvantage compresses throughout models (M1: $b = -1.192$; M4: $b = -0.149$, $p = .015$). After controlling for era, vote counts, and ratings, the commercial-only penalty is modest. LLMs do not disfavor commercial films categorically; the residual penalty for Set~C, after controlling for era and knowledge coverage, is largely accounted for by low IMDb ratings.

\textbf{Control measures.} Log IMDb votes contribute positively ($b = +0.094$, $p < .001$), affirming that training-data coverage predicts preference strength independently of set membership. IMDb user rating is the strongest single predictor ($b = +0.714$, $p < .001$) in the full model specification (M4). The Set~B coefficient attenuates when rating is added (M3: $b = +0.638$ $\to$ M4: $b = +0.554$), but remains large and significant, indicating that critical acclaim carries distinct evaluative signal. VIF diagnostics confirm collinearity is not harmful ($\mathrm{GVIF}^{1/(2\,\mathrm{df})}$: set = 1.49, era = 1.14, log IMDb votes = 1.61, IMDb rating = 1.55).

\textbf{Model comparison.} Sequential F-tests confirm significant improvement at each step: M1$\to$M2 ($F(3, 1594) = 36.53$, $p < .001$), M2$\to$M3 ($F(1, 1593) = 365.93$, $p < .001$), M3$\to$M4 ($F(1, 1592) = 467.98$, $p < .001$).  Era controls (M1$\to$M2) raise adjusted $R^2$ by 3.0 percentage points; adding log IMDb votes (M2$\to$M3) raises it by a further 10.5 points. The largest single-step gain occurs at M3$\to$M4 ($\textsuperscript{adj.}\ R^2 = 13.5$~pp), driven by the addition of IMDb user rating ($b = +0.714$, $SE = 0.033$). The fact that the popular acclaim signal adds more explanatory power than the coverage proxy indicates that training corpora encode evaluative valence (how films are rated) as well as how often they appear. The full model accounts for 54.0\% of the variance in pooled $\lambda_z$ scores ($\textsuperscript{adj.}\ R^2 = 0.540$).

\textbf{Robustness check: Additional measure of visibility and coverage.} Since IMDb vote counts index general audience engagement rather than visibility or token frequency in pretraining corpora directly, it may represent a noisy proxy. To address this issue, M3 was re-estimated including log Wikipedia revision count alongside log IMDb votes (M3\textsubscript{w}) (full results reported in the Appendix~D). Wikipedia is a documented component of major LLM pretraining corpora, so revision count provides a plausible proxy for sustained textual representation and editorial attention \cite{Brown_et_al_2020, Gao_et_al_2020}. Adding Wikipedia revisions alongside IMDb votes slightly improved model fit ($F(1, 1592) = 13.54$, $p < .001$). The Set~B coefficient changed from $b = +0.638$ to $b = +0.544$, indicating that the sign reversal is robust. However, due to the high correlation between two proxies ($r = 0.93$), Wikipedia revisions are excluded from M4 to avoid harmful collinearity ($\mathrm{GVIF}^{1/(2\,\mathrm{df})} = 3.06$ when entered together). Nevertheless, the convergence of results lends additional support to the coverage interpretation.

\section{Discussion}

The results demonstrate that LLM outputs exhibit stable evaluative orientations toward cultural objects. Across eight models from four families, critically acclaimed films consistently outperform commercially successful ones in forced-choice comparisons at rates that far exceed chance. The differences are not explained by differences in public visibility. The evaluative \textit{valence} of critical discourse, not its volume, is robustly associated with variations in these outcomes. The scale analysis reveals that critical acclaim orientation intensifies with model capability within every family. Model scale does not uniformly amplify baseline preference, though; increased scale along with the model's exposure to the specific films being evaluated likely lead to differential shifts across model families. Furthermore, we find a divergence between elicited preferences and general-audience recommendations, indicating that models maintain functionally distinct response modes. This suggests LLMs may not impose critical hierarchies in everyday use, yet does not eliminate the concern about downstream impacts. 

We elaborate these findings below. Throughout, we treat model orientation and elicited preference as behavioral regularities in observed output patterns. The results are consistent with, but do not establish, encoding at the representational level; the study does not make claims regarding internalized or cognitive dispositions.

\subsection{Critical Acclaim Orientation as a Robust, Cross-Model Property} % what happened

The consistency of critical acclaim orientation across all models, irrespective of family and capability tier, establishes that the pattern is not idiosyncratic to any single training regime. Evaluative hierarchies for films exist in LLM outputs, are reproducible under semantically equivalent prompt framings, and persist across models with different architectures and institutional origins.

A popularity or visibility explanation would predict the opposite. Commercially successful films generate far more online discourse than critically acclaimed yet obscure ones, and models that mirror the distributional density of their training data should favor commercial success. Our data, however, do not support this prediction. For all eight models, Set~B films win the majority of direct matchups against Set~C films despite having a fraction of online engagement (median 26,431 vs.\ 312,582 IMDb votes of Set~C).

The regression models untangle this relationship by decoupling a film’s visibility from its evaluative valence. Log IMDb votes---a proxy for a film's textual footprint across training corpora---contributes positively to preference strength (M3: $b = +0.235$; M4: $b = +0.094$). But public visibility and evaluative orientation are partially separable predictors. In unadjusted models, Set~B films score below Set~A (M1: $b = -0.219$), which means that dual-legitimacy films are preferred over critical-only ones. Once public visibility is controlled, this relationship reverses (M3: $b = +0.638$; M4: $b = +0.554$). The Set~C results tell a complementary story: the steep penalty initially suffered by commercial-only films compresses considerably across models (M1: $b = -1.192$; M4: $b = -0.150$), once IMDb user ratings are included.

These results imply that the \emph{visibility} of discourse surrounding a cultural object and the \emph{evaluative valence} of that discourse are distinct dimensions that jointly shape LLM outputs. The evaluative hierarchy reproduced by these models is consistent with the social stratification of cultural consumption. Large-scale survey data show that art-house and alternative cinema is concentrated among professional and university-educated audiences \cite{Bennett_et_al_2009}. Texts about Set~B films---reviews, canonical list entries, course syllabi, scholarly commentary---are likely produced by and for these same audiences. The findings are consistent with this discourse having shaped model outputs in ways that reproduce its evaluative logic. This interpretation extends Kozlowski et al.'s \citeyearpar{Kozlowski_et_al_2019} foundational work; whereas they recovered prestige dimensions from static word embeddings, we show that these entrenched hierarchies might surface in outputs when models are forced to make comparative evaluations of cultural works.

\subsection{Prestige Attraction, Commercial Discount, and Symbolic Exclusion} % how to interpret

The regression results complicate a simple prestige interpretation. The raw preference ordering in unadjusted models (M1 \& M2) follows a fairly intuitive hierarchy: Set~A (critical and commercial recognition) slightly above Set~B (critical only), which in turn dominates Set~C (commercial only). However, after accounting for public visibility and popular reception, this ordering reverses. This reversal is important because the remaining difference between Set~A and Set~B films is most likely commercial recognition.\footnote{An empirical challenge remains: it is, as far as we can tell, impossible to completely separate either of these attributes from each other or from other attributes of films that may provide alternative pathways to LLM evaluative preferences. We leave solutions to this issue to future work or until LLMs can better explain themselves.} After controlling for visibility and evaluative valence, the addition of commercial success is associated with lower estimated preference strength.

We consider three interpretations of this reversal. First, commercial success may contribute little additional evaluative lift once critical recognition is established---a \textit{prestige saturation} reading in which critical consecration is the binding constraint and commercial success adds nothing at the margin. Second, a \textit{training coverage} reading assumes that Set~B films are underrepresented in training corpora relative to their evaluative standing. The raw Set~B scores thus understate the true preference signal; once public visibility is controlled, their advantage over Set~A emerges. Third, on a more speculative note, commercial recognition may become mildly disadvantageous once evaluative and visibility-related factors are held constant. This is a \textit{commercial discount} reading in which model outputs are aligned not only with critical prestige, but also with distinctions internal to prestige itself, in which critical acclaim is valued more strongly when detached from commercial success.

This third possibility recalls Bourdieu's \citeyearpar{Bourdieu_1984} observation that ``Tastes (i.e., manifested preferences)'' are ``asserted purely negatively, by the refusal of other tastes.'' Bryson's \citeyearpar{Bryson_1996} analysis of musical dislikes extends this logic and shows that \textit{symbolic exclusion}, the targeted rejection of low-status cultural goods, operates as a mechanism of distinction alongside the endorsement of high-status ones. Viewed through this lens, critical acclaim orientation might comprise both attraction to critical acclaim and discounting of commercial success. This interpretation should be held cautiously, since the design does not measure anti-commercial discourse directly, and the coefficient pattern is also consistent with the other two readings.

 The Bradley--Terry framework we adopt here cannot distinguish whether a model is drawn toward a canonical film, repelled by its box-office counterpart, or both. These are asymmetric processes with arguably distinct origins in training data or possibly post-training stages of model development. A model may produce outputs consistent with high evaluative signal for a commercially obscure film it has encountered rarely, and simultaneously produce outputs consistent with low evaluative signal for a commercially successful film whose critical dismissals are densely represented across corpora. In a pairwise comparison, that asymmetry alone could produce a Set~B win. In short, the present design identifies relative selection behavior rather than positive attachment to critical prestige alone.

\subsection{Coverage, Scale, and Family-Specific Mechanisms} % why it happened

The scale effect---larger models within each family showing stronger critical acclaim orientation---is consistent across all four families, but the mechanism is not uniform. For OpenAI and Mistral, the B~vs.\ C win rate rises sharply with scale while the A~vs.\ C rate remains flat, a differential pattern consistent with the training coverage reading. Set~A films are well-known at all scales; Set~B films are critically acclaimed but commercially obscure, with median numbers of IMDb votes 47 times smaller than Set~A. As models scale, their representation of Set~B films plausibly becomes denser, and the preference signal for Set~B strengthens accordingly. The regression sign reversal corroborates this interpretation, indicating that coverage and evaluative orientation operate as partially separable factors in these families.

For Anthropic and Alibaba, by contrast, both B~vs.\ C and A~vs.\ C win rates rise at similar rates with scale. This parallel pattern cannot be explained by the coverage account: Set~A films are already well-known at smaller scales, so their improved win rates against Set~C at larger scales cannot be attributed to denser representation. Something else changes with scale for these families, but the current design cannot identify what. The two patterns are not mutually exclusive and both may be operating simultaneously in all families to varying degrees. The data establish that scale does not simply amplify a fixed orientation: the mechanism through which scale operates is family-dependent, and scale effects on elicited cultural preference should not be read as uniform evidence of deeper evaluative encoding without accounting for the confounding role of training data coverage, divergent post-training regimes, or other factors.

\subsection{Evaluation, Recommendation, and Contextual Activation} %why it matters

What downstream impacts might critical acclaim orientation create? The divergence between evaluative and general audience recommendation prompt framings---mean $\rho \approx 0.26$--$0.31$ between the two framing types, versus $\rho \approx 0.85$--$0.88$ within evaluative prompts alone (``prefer'', ``like'', ``taste'')---illustrates that models produce distinct ranking patterns depending on how the elicitation is framed. If models can already distinguish ``what to evaluate highly'' from ``what to recommend to a general audience,'' then we would not expect critical acclaim orientation to drive biased recommendation feedback loops.

A more precise concern relates to implicit framing and contextual activation. Many real-world activities of LLMs engaged in cultural discernment---summarizing films, suggesting films to watch, describing what is worth seeing, ranking options---may not explicitly signal whether evaluation or recommendation is requested. The model must infer the appropriate mode from context, and that inference is not always transparent. When the framing is ambiguous or subtly evaluative, outputs consistent with critical acclaim orientation may surface even where user intent is closer to audience-fit recommendation. In addition, model behavior varies substantially depending on user history and metadata \cite{Wang_et_al_2025}, introducing further variability in which orientation may be activated. Models may not always impose critical hierarchies in recommendation contexts, but implicit framing of queries or inferential reasoning based on user interaction traces may activate critical acclaim orientation in subtle ways. Model users and auditors alike may not be able to anticipate reliably which cultural hierarchy a given interaction reproduces.

\subsection{Cultural Hierarchies within Dominant Culture} %theoretical implicaiton
One interpretation of these findings concerns the structure of cultural dominance itself. If critical acclaim orientation reflected only public visibility, existing accounts of WEIRD overrepresentation in training corpora would largely explain the result \cite{Atari_et_al_2023}. But commercially successful films are themselves disproportionately products of WEIRD cultural production and consumption. The observed distinction therefore appears less consistent with global representational inequality than with competition between evaluative logics internal to dominant cultural strata. The relevant distinction may be between ``lowbrow,'' mass cultural consumption and the credentialed taste-making practices of ``high'' cultural elites. Whether this reflects the evaluative logic of critical discourse incorporated during pretraining, the preferences of culturally positioned annotators encoded through reinforcement learning from human feedback, or some combination remains an open question---and one with implications for how the social position of those who produce and curate training data propagates into model outputs at scale.

\subsection{Limitations and Future Directions}

A number of limitations bound our findings and interpretation. First, this study examines a single cultural domain and evaluates a finite set of models. The question of whether critical acclaim orientation generalizes to music, literature, or visual art remains unknown, as do other dimensions of generalization beyond the results we report here. 

Secondly, the elicitation design constrains the stability of the results and the interpretability of the mechanisms behind them. The preference elicitation prompt uses a film title and release year as stimuli, a compressed representation that may activate whatever associations the model has formed around those tokens. Whether results would change if alternative or richer stimuli were provided (e.g., plot summaries, metadata) is untested. The elicitation design also cannot support definitive insights into the generative processes that produce model outputs. Admittedly, this is appropriate to the primary target of the current study: understanding preferences elicited directly from publicly available LLMs. Future work might investigate reasoning traces, mechanistic activation patterns, or related approaches to better understand the sources and mechanisms of critical acclaim orientation. Whether models are retrieving cached evaluative associations, computing comparisons from distributed representations, or instantiating some other process is not addressable with the current design. 

Third, all comparisons were conducted in English. Film titles were presented following the conventions of the source corpora (TSPDT and BOM), which are rooted in Anglophone critical discourse. Some non-English films were given translated titles (e.g., \textit{The Rules of the Game}), some retained their original titles (e.g., \textit{Au hasard Balthazar}), and others appeared in Romanized form (e.g., \textit{Rashomon}). These conventions are culturally familiar within Anglophone film criticism but may not be the most natural identifiers for models. It is untested whether the results would differ if prompts were issued in other languages \cite{Lu_et_al_2025}, or if the film titles were provided in their original scripts or language-specific conventions. In a similar vein, the corpus's critical canon is drawn from an English-language aggregation (TSPDT) that leans toward Western European and North American film traditions; preference patterns within non-Western critical canons may differ in ways that would produce different comparison sets and sample attributes.

Several natural extensions of this work include: (1) replication across cultural domains such as music and literature, with analogous critical-commercial distinctions indexed by established external sources. (2) Anonymized or partial-metadata conditions, testing if critical acclaim orientation persists when film titles and release years are withheld, or when plot summaries are provided. (3) Multilingual elicitation testing whether results vary when prompts are issued in the dominant language of each model family's training data. (4) Extension to non-Western critical canons, probing whether the orientation reflects universal patterns of prestige discourse or specifically Western critical traditions likely embedded disproportionately in training data.

\section{Conclusion}

To our knowledge, this study provides the first systematic measurement of LLM cultural preference across models, families, and capability tiers, using a methodology that is reproducible. The finding that critical acclaim orientation is robust across eight models from four institutional origins, intensifies with scale, and is not reducible to public visibility suggests that the evaluative hierarchies embedded in critical discourse are a structural feature of how frontier models engage with cultural objects.

Beyond the immediate findings on film evaluation, this study raises a broader question about how cultural values are studied and audited in AI systems. Multiple dimensions of model bias and unfairness are well established. The results introduce a concern that models may also reproduce hierarchies among cultural objects. Since aesthetic and cultural domains lack an objective baseline against which outputs can be neatly calibrated, any attempt to align model behavior inescapably involves a normative choice about which evaluative standard to prioritize. For applications that mediate recommendation and cultural discovery, transparency regarding these priorities is indispensable.

\section{Acknowledgments}

The authors thank OpenAI for providing API credits through their research access program, which supported a portion of the model evaluations reported in this study.

\bibliography{aaai2026}

\end{document}